\pdfoutput=1
\def\year{2022}\relax
\documentclass[letterpaper]{article} % DO NOT CHANGE THIS
\usepackage{aaai22}  % DO NOT CHANGE THIS
\usepackage{times}  % DO NOT CHANGE THIS
\usepackage{helvet}  % DO NOT CHANGE THIS
\usepackage{courier}  % DO NOT CHANGE THIS
\usepackage[hyphens]{url}  % DO NOT CHANGE THIS
\usepackage{graphicx} % DO NOT CHANGE THIS
\usepackage{natbib}  % DO NOT CHANGE THIS AND DO NOT ADD ANY OPTIONS TO IT
\usepackage{caption} % DO NOT CHANGE THIS AND DO NOT ADD ANY OPTIONS TO IT
\DeclareCaptionStyle{ruled}{labelfont=normalfont,labelsep=colon,strut=off} % DO NOT CHANGE THIS
\usepackage{algorithm}
\usepackage{algorithmic}

\usepackage{newfloat}
\usepackage{listings}
\floatstyle{ruled}
\newfloat{listing}{tb}{lst}{}
\floatname{listing}{Listing}
\usepackage{kotex} % Korean %added package
\usepackage{multirow}
\usepackage{diagbox}
\usepackage[caption=false]{subfig}
\usepackage{booktabs}
\usepackage{amsmath}
\usepackage{amssymb}

\title{Adversarial Attack for Asynchronous Event-based Data}
\author{
    Wooju Lee\textsuperscript{\rm 1}
    and Hyun Myung\textsuperscript{\rm 1}\thanks{The corresponding author.}\\
}
\affiliations{
    \textsuperscript{\rm 1}Urban Robotics Lab, School of Electrical Engineering, \\Korea Advanced Institute of Science and Technology, Republic of Korea\\
    
    dnwn24@kaist.ac.kr, hmyung@kaist.ac.kr
}

\usepackage{bibentry}
\begin{document}

\maketitle

\begin{abstract}
Deep neural networks (DNNs) are vulnerable to adversarial examples that are carefully designed to cause the deep learning model to make mistakes. Adversarial examples of 2D images and 3D point clouds have been extensively studied, but studies on event-based data are limited. Event-based data can be an alternative to a 2D image under high-speed movements, such as autonomous driving. However, the given adversarial events make the current deep learning model vulnerable to safety issues. In this work, we generate adversarial examples and then train the robust models for event-based data, for the first time. Our algorithm shifts the time of the original events and generates additional adversarial events. Additional adversarial events are generated in two stages. First, null events are added to the event-based data to generate additional adversarial events. The perturbation size can be controlled with the number of null events. Second, the location and time of additional adversarial events are set to mislead DNNs in a gradient-based attack. Our algorithm achieves an attack success rate of 97.95\% on the N-Caltech101 dataset. Furthermore, the adversarial training model improves robustness on the adversarial event data compared to the original model.
\end{abstract}
\begin{figure}[ht]
\centering
\includegraphics[width=0.9\columnwidth]{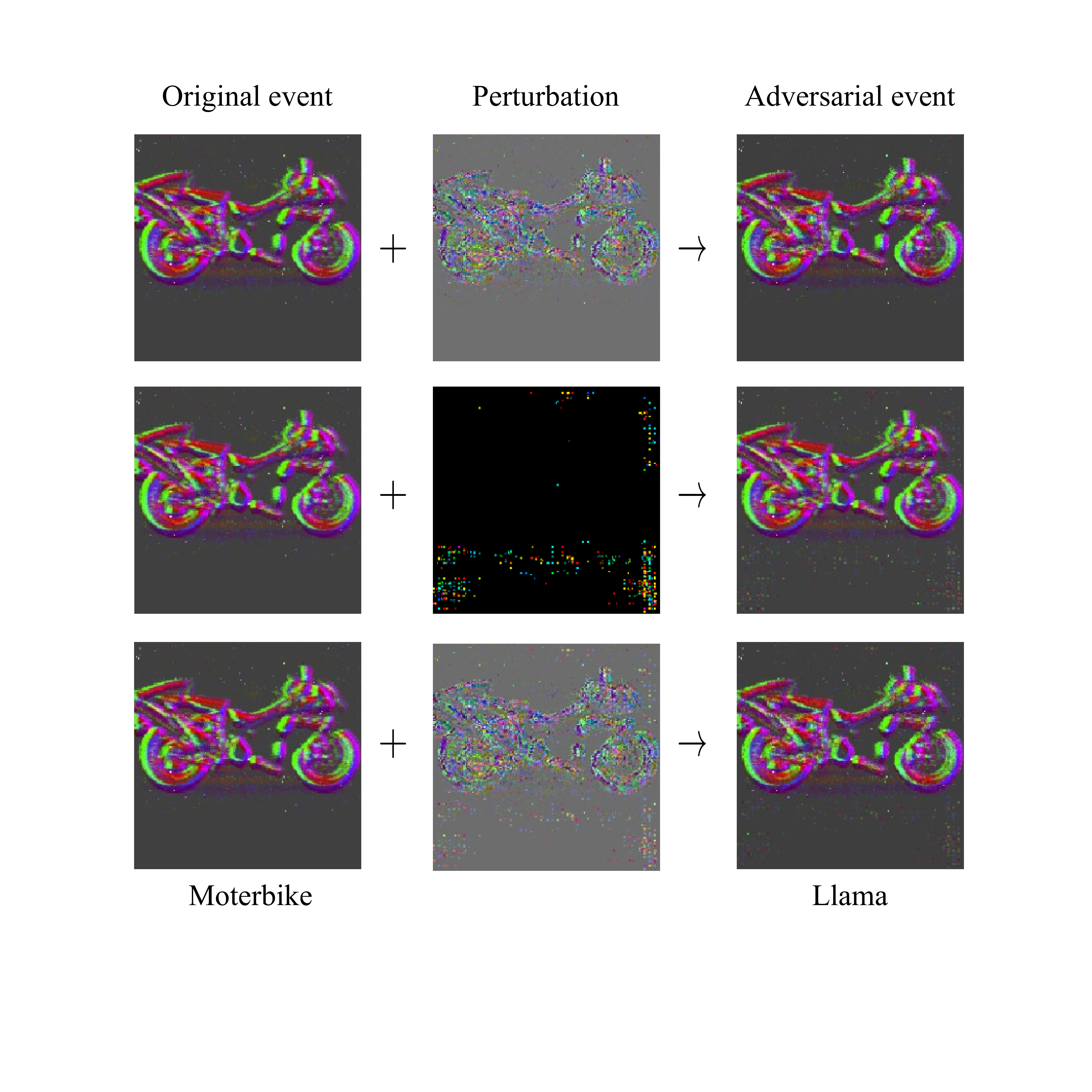}
\caption{Grid representations of adversarial events. The event times are separated by three channels and converted to 2D RGB images. Shifting events (top). Generating additional adversarial events (middle). A combined method (bottom).}
\label{fig1}
\end{figure}
\section{Introduction}
Although deep neural networks (DNNs) have achieved great success in various domains, DNNs are vulnerable to adversarial examples. Adversarial examples can mislead DNNs by adding imperceivable perturbations to the original data. They raise security concerns for DNNs in safety-critical scenarios, such as face recognition and autonomous driving, etc. For secure and robust deep learning, adversarial attacks have been extensively studied for data such as 2D images \cite{c:a2, c:a3, c:a7}, 3D point clouds \cite{c:a4, c:a5}, and natural languages \cite{c:a8, c:a9}. However, studies on event-based data are limited.

An event camera is an asynchronous visual sensor with high dynamic range and temporal resolution. Instead of capturing the brightness of images at a fixed rate, an event camera measures brightness changes (called events) for each pixel independently. An event camera  can be an alternative to a traditional camera in situations that require high-speed and high-dynamic range, such as autonomous driving and unmanned aerial vehicles (UAVs) \cite{c:e1, c:e7}. Although event-based deep learning models have been studied in various fields such as object recognition \cite{c:e4, c:e5, r:e8}, gesture recognition \cite{c:e6}, and optical flow \cite{c:e2, c:e3}, adversarial attacks on the event-based deep learning model raise security problems. In a recent study, Marchisio et al. \shortcite{c:a6} generated 2D adversarial images in event-based data. They projected event-based data to 2D images and generated 2D adversarial images. However, event-based data cannot be retrieved from 2D adversarial images. As the input data of an event camera cannot be attacked, it does not pose a real threat to the event-based deep learning. In this paper, we study how to generate adversarial examples for raw event-based data and train the robust deep learning model against adversarial examples.

As to the attacking target, we focus on the event spike tensor (EST) representation \cite{c:e5} of event-based data, which is a generalized grid representation of the event-based data. By projecting event-based data over the time or polarity axis, other types of grid representation can be derived from EST. Here,  we demonstrate the adversarial attack algorithm by conducting experiments on the EST and the projected EST.

There are challenges of generating adversarial examples in the conventional way due to the characteristics of event-based data.
First, the adversarial attack performance is affected by the frequency of relative motion between an event camera and objects. An event camera captures static objects with repeated movements of the camera. When the frequency of the relative motion is high, a perturbation can make some events mis-regarded as original events rather than adversarial events.
Second, the success rate of an attack depends on the size of event-based data. The size of event-based data is not fixed because an event is measured whenever a brightness changes. When the event-based data is small, there are fewer attacking targets. The number of attacking targets can limit the attack performance. 

In this paper, we create adversarial examples that consider the characteristics of the aforementioned event-based data. Our algorithm shifts the time of the original events and generates additional adversarial events. When shifting the time of the original events, we set the optimal perturbation size of an attacker to perturb the event-based deep learning model. The perturbation size depends on temporal scales in the grid representation and frequency of camera motion. Generation of additional adversarial events is proceeded with two steps. First, we add null events to the event-based data to generate additional adversarial events. A null event is an empty space to generate additional adversarial events and to control the perturbation of the adversarial attacks. By adding null events to the original event, additional adversarial attacks can be performed at any time and space. Second, we set the time and location of additional adversarial events in a gradient-based attack. We set the location of additional adversarial events based on the loss gradient of the null events. Then, we determine the time of additional adversarial events based on a projected gradient descent (PGD) attack algorithm \cite{c:a3}.

Our algorithm shows a 97.95\% success rate for untargeted attack and 70.68\% success rate for random targeted attack on the N-Caltech101 dataset \cite{r:d1}. We utilize the proposed attack algorithm to train networks that perform well both on the original events and adversarial events. Furthermore, we discuss the transferability of adversarial events between representation models. The contributions of this paper can be summarized as follows.
\begin{itemize}
\item We propose an adversarial attack algorithm for event-based deep learning model for the first time.
\item An optimal perturbation size is set to attack the time of the original events.
\item Null-events make it possible to generate additional adversarial events at any time and space.
\item Additional adversarial events are generated in a gradient-based attack.
\item We generate adversarial events for various types of grid representation and kernel function.Through experiments on grid representations and kernel functions, we prove that our method can generate adversarial events for grid representation model in general .
% \item We find robust grid representations of event-based data against adversarial examples.
\item We train event-based deep learning models that are robust to adversarial examples.
\end{itemize}
\section{Related Work} 
\subsection{Event-based Data}
Due to the sparse and asynchronous characteristics of the event-based data, typical event-based algorithms aggregate events into a grid representation. Event-based data consists of a stream of events that encode the location, time, and polarity of the brightness changes \cite{c:e10}. When event-based data is converted into a grid representation, $(x,y)$ position coordinates become pixel position coordinates, and time coordinates become tensor values. As each event alone contains little information, events must be aggregated into a grid representation. The event count model \cite{c:e1, c:e2} measures event counts for each pixel and polarity, but it discards temporal and polarity information. The two-channel model \cite{c:e1,c:e2,c:e4} uses only one time as a tensor value. The voxel grid model \cite{c:e3}  uses all the temporal information, but it discards polar information. The EST model \cite{c:e5} processes events with a specific kernel and converts event-based data into a grid representation of multiple channels. Compared to other representations, EST keeps the most information from raw event streams. EST is a generalized grid representation, thus it can be transformed to other representations by projecting over the temporal or polar axis. For example, projecting the EST over the temporal axis leads to the two-channel, and projecting the EST over the polar axis leads to the voxel grid. We demonstrate our attacking algorithm for grid representation of event-based data with the EST and the projected EST. 
\subsection{Adversarial Attack}
Szegedy et al. \shortcite{c:a1} first pointed out that DNNs were vulnerable to intentionally designed adversarial examples. Adversarial examples can easily fool DNNs by adding imperceptable perturbations to the original data. Since 2014, several approaches \cite{c:a2, c:a3, c:a10, c:a11, c:a12} have been proposed to generate adversarial examples to attack models effectively. Fast gradient sign method (FGSM) \cite{c:a2} is one of the most popular one-step gradient-based approaches for $L_{\infty}$-bounded attacks. A PGD attack \cite{c:a3} iteratively applies FGSM multiple times, and is one of the strongest adversarial attacks. But, they mainly target 2D images. Recent studies have shown that DNNs are vulnerable to 3D adversarial objects. Xiang et al. \shortcite{c:a4} generated 3D adversarial point clouds through point perturbation or point generation. Tsai et al. \shortcite{c:a5} generated physical 3D adversarial objects and proved that physical objects can mislead the DNNs to make wrong predictions. While adversarial examples of 2D images and 3D point clouds have been extensively studied, studies on event-based data are limited. 
\subsection{Adversarial Attack on the Event-based Data}
Marchisio et al. \shortcite{c:a6} studied various types of adversarial examples for event-based data. 
% They transformed the original event to 2D images and performed a typical adversarial attack on the 2D images. The 2D images are intermediate features of event-based deep learning model. 
They performed a typical adversarial attack on intermediate features of the event-based deep learning model. The intermediate features are 2D grid representations of event-based data. However, adversarial examples of the input event cannot be retrieved from 2D adversarial features. Inverse operation is not possible, because input events are convolved with nonlinear kernel functions and aggregated into grid representations. As the input data of an event camera cannot be attacked, it does not pose a real threat to the event-based deep learning. To the best of our knowledge, this is the first study to generate adversarial examples for event-based data. 
% And an adversarial attack on the grid representation easily generates events even with small changes of tensor value. In the grid representation model for event-based data, time value 0 is given as a tensor value to the $(x,y)$ coordinate position where an event does not exist. As event-based data is sparse, most pixels have a time value of 0 in the grid representation. Therefore, adversarial attack on the grid representation can trigger events, resulting in a noticeable change of data. 
\section{Method}
\subsection{Event-based Deep Learning}
\subsubsection{Event-based Data}
An event camera has independent pixels that trigger events when there is a change in log brightness:
\[L(x,y,t)-L(x,y,t-\kappa)\geq pC,\]
where $p \in \{ -1, +1\}$ is the polarity of the brightness change; $C$ is the contrast threshold; and $\kappa$ is the elapsed time after the last event occurence at the same pixel. Events are triggered by brightness changes and relative motion between the event camera and objects. The event camera captures static objects with the repeated movements of the camera. As the motor is moving repeatedly, event-based data includes recurrent information. In a given time interval $\tau$, triggered events are defined as point-sets:
\[\mathcal{E}=\{e_k\}_{k=1}^N=\{(x_k,y_k,t_k,p_k)\}_{k=1}^{N},\]
where $x_k \in [0, W]$ and $y_k \in [0, H]$ are the event's spatial coordinates; $p_k \in \{-1, +1\}$ is the event polarity; and $t_k \in (0, 1]$ is the event's normalized timestamp. Data is not defined in any location without an event.
\begin{figure}[t]
\centering
\includegraphics[width=0.9\columnwidth]{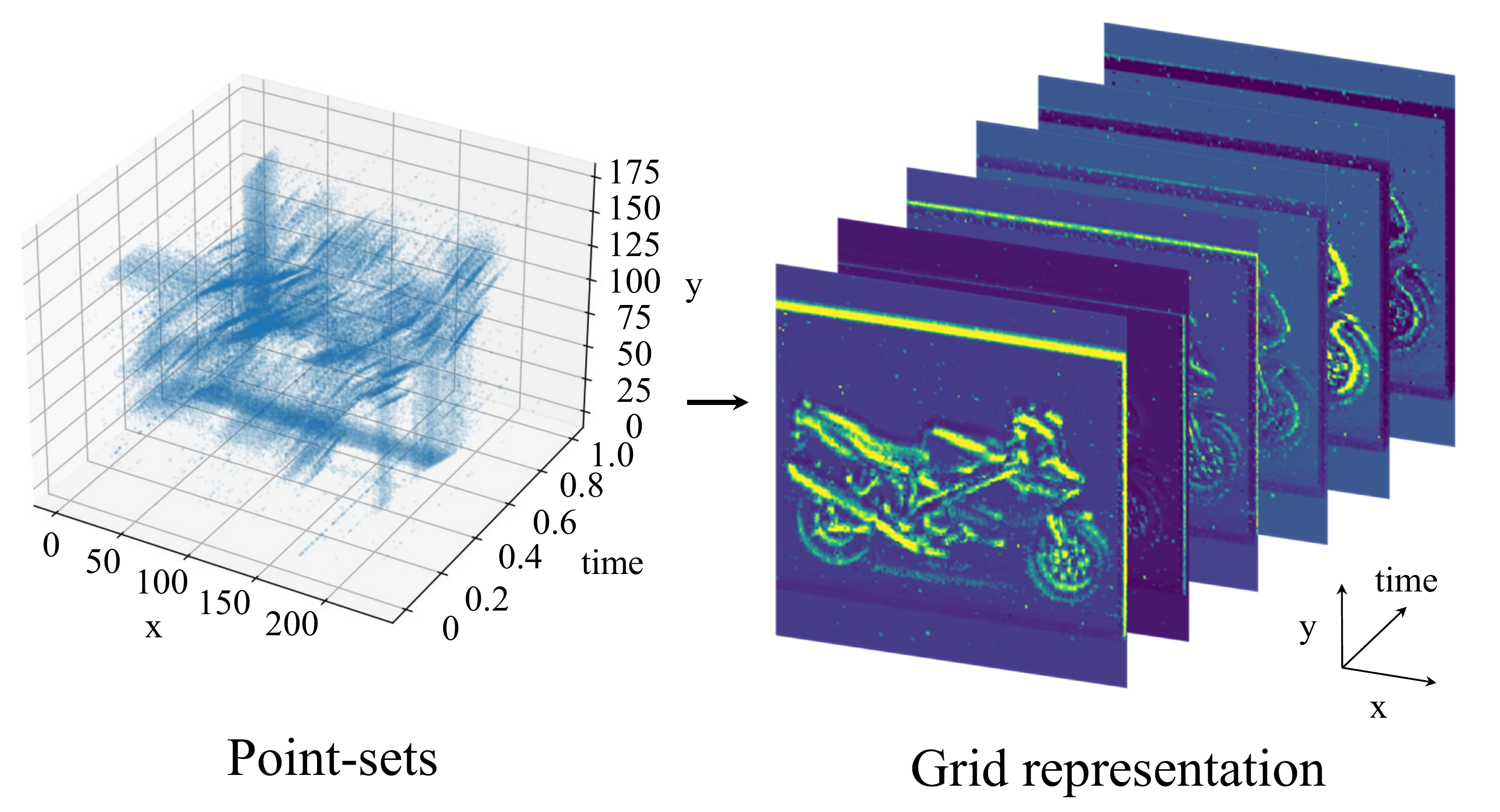} 
\caption{Point-sets (left) and grid representation (right) of event-based data. Point-sets of event-based data are mapped to tensor map $T \in \mathbb{R}^{C\times W \times H}$. The event times are separated by three temporal bins. The temporal bins of + polarity and - polarity are concatenated along the temporal dimension.}
\label{fig2}
\end{figure}
\subsubsection{Kernel Convolution}
Because event-based data contains numerous events, each separate event has limited information. Use of a suitable kernel derives a meaningful signal from events with high temporal resolution. The EST representation model convolves the event-based data with a multilayer perceptron (MLP) kernel, which is a learnable kernel, or a trilinear voting kernel $k(x,y,t)=\delta(x,y)\max (0, 1-|\frac{t}{\tau}|)$. An exponential kernel, $k(x,y,t)=\delta (x,y)\frac{1}{\tau}\exp(-\frac{t}{\tau})$, is used to construct spatio-temporal features of HOTS \cite{r:e8} and HATS \cite{c:e4}. Each time value of the event-based data is convolved with appropriate number of temporal channels by a suitable kernel  for grid representation. 
% \[\sum_{e_k\in \mathcal{E}} k(x,y,t-t_k).\]
\begin{figure*}[t]
\centering
\includegraphics[width=0.95\textwidth]{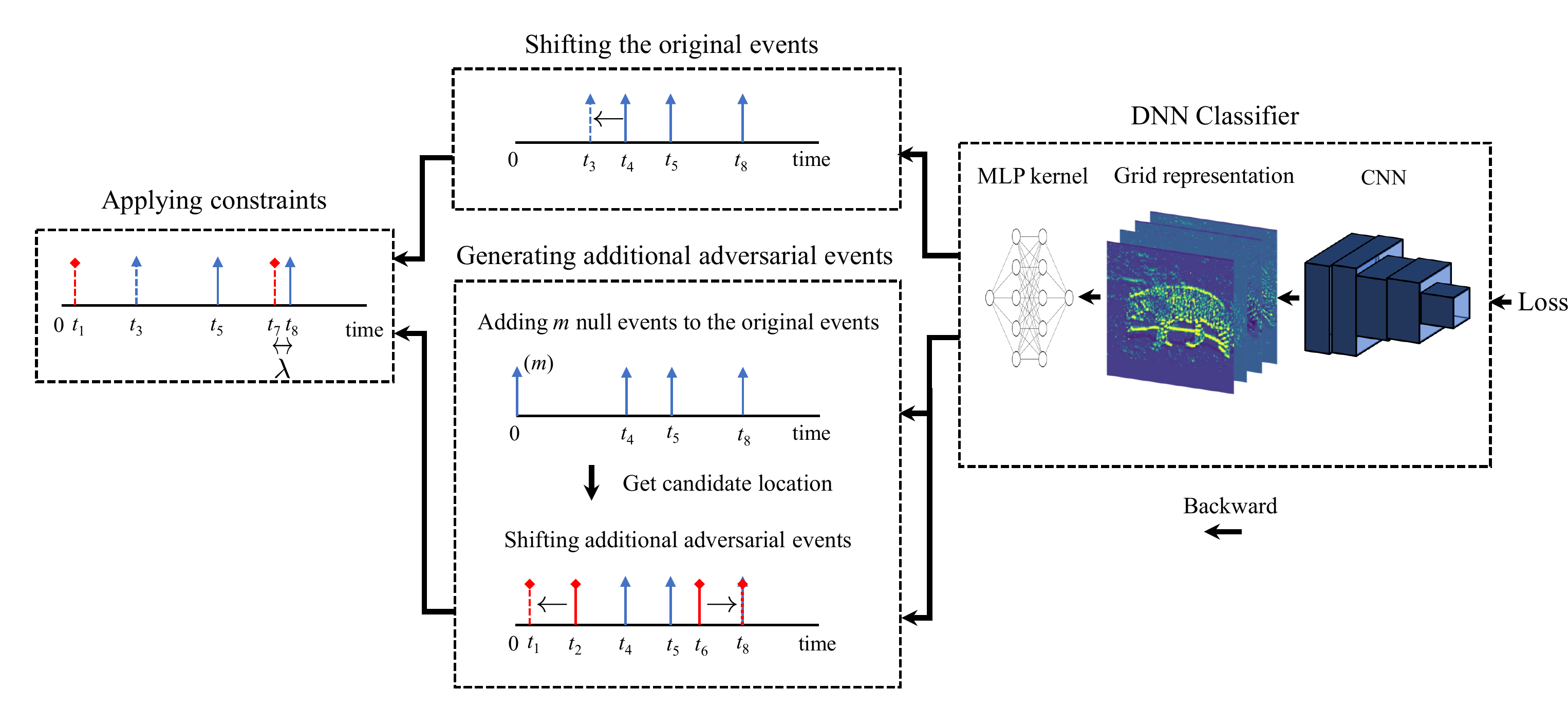} % Reduce the figure size so that it is slightly narrower than the column.
\caption{Shifting the original events and generating additional adversarial events. The original events are shifted with perturbations relative to the temporal scale. Solid arrows and dotted arrows indicate the original events and shifted events, respectively. Null events are generated in the location without any event and concatenated to the original events. The number in parentheses means the number of redundant events. The location of additional adversarial events are set based on the loss gradient of the null event. A line with a diamond head indicates an additional adversarial event. A PGD attack algorithm is performed on the original events and additional events. The redundant events are separated by a minimum time resolution $\lambda$. }
\label{fig3}
\end{figure*}
\subsubsection{Grid Representation}
As event-based data includes recurrent information, an event-based vision algorithm divides event-based data into the temporal bins for efficient processing. Given the convolved signals, the EST representation model computes a tensor map $T \in \mathbb{R}^{C\times W \times H}$, where $W$ is the width; $H$ is the height; and $C$ is the number of channels. The spatio-temporal coordinates $(x_k, y_k, t_k)$ lie on a 3D voxel grid, i.e., $x_k \in \{0, 1, 2, \cdots, W-1\}, y_k \in \{0, 1, 2, \cdots, H-1\}$, and $t_k \in \{0, \Delta t, \cdots, B\Delta t\}$, where $\Delta t$ is the temporal scale for each temporal bin and $B$ is the number of temporal bins \cite{c:e5}. The convolved time values are assigned to each temporal bin and used as the tensor value of the tensor map $T$. The temporal bins of + polarity and - polarity are concatenated along the temporal dimension. The temporal channel $C$ of the EST is $2\cdot B$. The temporal scale $\Delta t$ is $1/B$ as $t_k \in (0, 1]$ is the normalized timestamp.

In the current event-based grid representation model, time 0 is given as a tensor value for the $(x,y)$ coordinate position without any event. Mapping time 0 to the nonexisting event preserves the structure and information of the events. After the event-based data is converted to the grid representation, data is fed into the convolutional neural network (CNN).
\subsection{Shifting Original Events}
We first recall the PGD attack to generate adversarial examples. A PGD attack is a typical multistep attack algorithm for an $L_{\infty}$ bounded adversary. Given the 2D image $u \in \mathbb{R}^{C\times H \times W}$, the PGD attack generates an adversarial example as follows:
\[
u^{i+1} = \text{Proj}_{\epsilon}\{u^i+\alpha \cdot sign[\nabla_{u}J(\theta, u^i, y)], \}
\]
where $i$ the iteration of the attack; $\epsilon$ the perturbation size; $\text{Proj}_{\epsilon}$ projects the adversarial sample into the $\epsilon-L_{\infty}$ neighbor of the benign sample; $\alpha$ is the step size; $J$ the loss function; $\theta$ parameters of the DNNs; and $y$ the true label.

The event-based deep learning model extracts features based on the event times. Therefore, we add imperceptible perturbations to the original event times and intentionally mislead the DNNs. The original event times are attacked by a PGD attack as follows:
\[
t^{i+1} = \text{Proj}_{\frac{\epsilon}{B \cdot f}}\{t^i+\frac{\alpha}{B\cdot f} \cdot sign[\nabla_{t}J(\theta, t^i, y)],\}
\]
where $t$ is the timestamp of $\mathcal{E}$, $B$ is the number of temporal bins, and $f$ is the relative frequency. We divide the perturbation size $\epsilon$ and step size $\alpha$ by the number of temporal bins $B$ and relative frequency $f$. For event-based data, convolved signals are given to each temporal bin of grid representation. As $B$ increases, the temporal scale of each bin decreases. Therefore, we use the relative perturbation size $\frac{\epsilon}{B \cdot f}$ and step size $\frac{\alpha}{B \cdot f}$ to perturb the time value in the smaller time scale $\Delta t$. 

\begin{figure}[t]
\centering
\includegraphics[width=0.75\columnwidth]{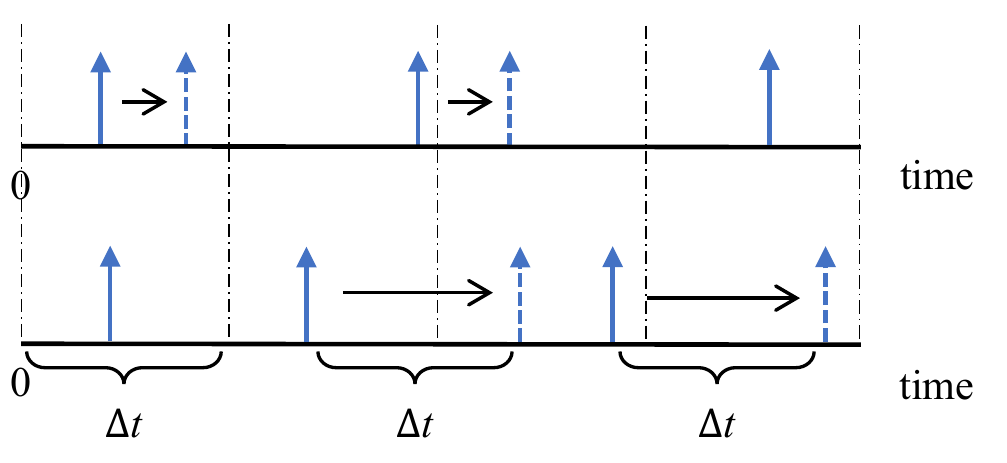} 
\caption{Effect of temporal scale $\Delta t$ to the shifting events. The entire time is divided by 4 temporal bins. Solid arrows and dotted arrows indicate the original events and shifted events, respectively. Perturbations less than temporal scale $\Delta t$ (top). Large perturbation can shift the events to the same location of another temporal bin (bottom). }
\label{fig4}
\end{figure}

The perturbation size is generally proportional to the attack success rate because more perturbation can mislead DNNs easily. However, larger perturbation size can shift the events to the same location of another temporal bin. In the bottom of Figure \ref{fig4}, large perturbation can shift the event to the same location of another temporal bin. As another temporal bin may contain the same information, the attack performance is reduced. Therefore, the relative perturbation size $\frac{\epsilon}{B \cdot f}$ balances the two opposing cases. We demonstrate it in the experiment section.

An adversarial attack on the original events only deals with $N$ time values. However, an infinite number of adversarial events can be theoretically generated in the region where $x\in\{0,1,..W\}, y\in\{{0,1,..H}\}$, and normalized timestamp $t\in (0, 1]$. In a pixel with $k$ events, only $k$ events can be attacked and additional adversarial events cannot be generated. In addition, a pixel without an event cannot have any adversarial event.

\subsection{Null Event}
\begin{figure}[t]
    \centering
    \subfloat[]{\includegraphics[width=0.8in]{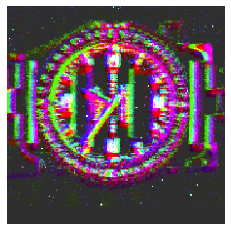}} 
    \subfloat[]{\includegraphics[width=0.8in]{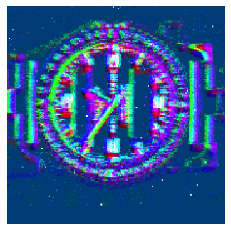}}
    \subfloat[]{\includegraphics[width=0.8in]{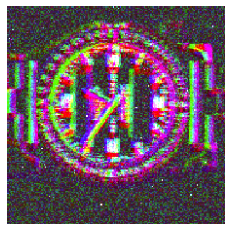}}
    \subfloat[]{\includegraphics[width=0.8in]{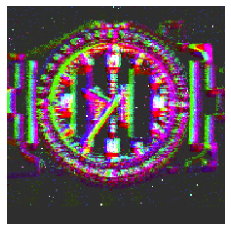}}
    \caption[]{Effect of null events with respect to the time value. (a) is the original event data. (b), (c), and (d) contains null events. (b) contains the null event of time value $-1$. (c) contains the null event of random time value. (d) contains the null event of time value $0$. (d) is the same as the original data.}
    \label{effect of_null_event}
    \label{fig5}
\end{figure}

Therefore, we propose to add null events to the original event-based data. The time value of the null event is set to 0 to satisfy two conditions. The first condition is that the null event should be distinguished from the original event. The normalized time value of the original event lies in the interval $(0, 1]$. Because the time value $t$ is recorded when the brightness changes, there is no original event at time 0. The time value of the null event should be set to the interval excluding $(0, 1]$. Second, the null event should not affect the prediction of DNNs for the original event. A nonzero time value of the event is mapped to a nonzero tensor value in the grid representation. Even a small time value can make real events in the grid representation. Therefore, we set the time value of the null event to 0. As shown in Figure \ref{fig5}, null event of time value 0 does not change the original events.

When null events are added to the original event, adversarial attacks can be performed at the desired location and time. Furthermore, a null event controls the perturbation size. For adversarial attacks on 2D images, perturbations of pixel values are limited to make invisible changes to the human eye. Similarly, adversarial attacks on event-based data can achieve the same effect by adjusting the number of null events. The effect of null events is included in the supplemental material.

An original event does not have an event with time value 0, but a null event is defined as a multiset that allows redundancy at time value 0. $m$ represents the number of null events added to each pixel. Null events can be added differently for each pixel, but in this paper, we add the same number of null events to the location without any event for simplicity as follows:
\[\mathcal{N}^m=\{(x_l,y_l,0,p_l)^m\},\] 
where no event exists in $(x_l,y_l, p_l)$. Null events $\mathcal{N}^m$ are added to the original events to make adversarial events $\mathcal{E}^{'}$. 
% \[\mathcal{E}^{'} \leftarrow \text{CONCAT}(\mathcal{E}, \mathcal{N}^m)\] 
\subsection{Generating Additional Adversarial Events}
By adding null events, adversarial events can be created in any space and time. However, additional adversarial events should be generated at the appropriate location and time to attack DNNs. The method proceeds in two stages. First, the location of additional adversarial events are set based on the null event's loss gradient. An event is generated when the time value of a null event increases from 0 to the time interval $(0, 1]$. If the generated event increases the loss, the generated event is assumed to have attacked the original event properly. Therefore, the position of loss gradient's + component indicates the location where the original event should be attacked. The null events' loss gradients are defined as $\nabla_{t_l}J(\theta, t_l, y)$, where $t_l \in (0, 1]$; $\theta$ is the model parameter; and $y$ is the true label. We use the location $(x_l, y_l, p_l)$ of the top $p\%$ loss gradients to generate additional adversarial events. Then, we set the time of additional adversarial events. We initialized additional adversarial events as random values between 0 and 1. We use the PGD attack for the time perturbation of additional adversarial events. The overall procedure for generating additional adversarial events is explained in Algorithm \ref{alg:algorithm}.
\begin{algorithm}[t]
\caption{Generating Additional Adversarial Events}
\label{alg:algorithm}
% \textbf{Input}: Original event $\mathcal{E}=\{(x_k,y_k,t_k,p_k)\}_{k=1}^{N}$
\textbf{Input}: Original event $\mathcal{E}=\{(x_k,y_k,t_k,p_k)\}_{k=1}^{N}$
\\
\textbf{Parameter}: Null events per grid $m$, iterations $I$, step size $\alpha$, perturbation size $\epsilon$, true label $y$  \\
\textbf{Output}: Adversarial event $\mathcal{E}_{a}$ % $\mathcal{E}^{'}=\{(x_k,y_k,t_k,p_k)\}_{k=1}^{N(1+m)}$
\begin{algorithmic}[1] %[1] enables line numbers
\STATE Get the location $\{(x_l, y_l, p_l)\}$ without any event from $\mathcal{E}$
\STATE Initialize null event $\mathcal{N}^m$ \\
$\mathcal{N}^m \leftarrow \{(x_l,y_l,t_l,p_l)^m\}$, $\forall t_l = 0$
\STATE $\mathcal{E}^{'} \leftarrow \text{CONCAT}(\mathcal{E}, \mathcal{N}^m)$   
\STATE $(x_l, y_l, p_l) \leftarrow$ Select the top $p\%$ of $\mathcal{N}^{m}$ with $\nabla_{t_l}J(\mathcal{E}^{'}, y, \theta)$
\STATE Initialize additional adversarial events $\mathcal{N}^{'}$ \\
$t_l \leftarrow$ rand$(0, 1]$ \\
$\mathcal{N}^{'} \leftarrow \{(x_l,y_l,t_l,p_l)\}$
\STATE $\mathcal{E}_{a} \leftarrow \text{CONCAT}(\mathcal{E}, \mathcal{N}^{'})$  
\FOR{$i=1,2, ... I$}
% \STATE $t_l \leftarrow Proj(t_l + \alpha \cdot sign(\nabla_{t_l}L(\mathcal{E}_{a}, y, \theta)))$
\STATE $t_l \leftarrow \text{Proj}_{\epsilon}(t_l + \alpha \cdot sign(\nabla_{t_l}J(\mathcal{E}_{a}, y, \theta)))$
% \STATE $t_l \leftarrow \test{clip}(t_l, 0, 1)$
% \STATE $t_k \leftarrow \test{clip}(t_k, 0, 1)$
\ENDFOR
\STATE $\text{Return }\mathcal{E}_{a}$
\end{algorithmic}
\end{algorithm}
\subsection{Constraint of Event-based Data}
After generating the adversarial events, we apply constraints on the adversarial events because events cannot occur simultaneously at the same location. Thus, when adversarial events are generated simultaneously at the same location, the adversarial events are separated by a minimum time resolution $\lambda$. In addition, the time values of adversarial events are limited in the interval (0, 1].
\begin{table*}[t]
\centering
%\resizebox{.95\columnwidth}{!}{
\begin{tabular}[t]{c|c|c|cccc}
    \toprule
    Attack & Representation & Kernel & Shifting & Generating & Generating and shifting  \\
    \midrule
    \multirow{7}{*}{Untargeted attack} 
    & EST (10) & MLP & 95.50 & 14.65 & \textbf{95.87} \\
    & EST (5) & MLP & 96.13 & 24.59 & \textbf{100.0} \\
    & EST (1) & MLP & 95.96 & 61.47 & \textbf{99.0} \\
    & Voxel grid (10) & MLP & 95.14 & 12.15 & \textbf{96.12} \\
    & Two-channel (1) & MLP & 97.74 & 63.35 & \textbf{99.42} \\
    & EST (10) & Trilinear & 94.16 & 14.62 & \textbf{96.03} \\
    & Two-channel (1) & Trilinear & 97.02 & 63.43 & \textbf{99.21}  \\
    \midrule
    \multirow{7}{*}{Random targeted attack} 
    & EST (10) & MLP & 41.64 & 29.69 & \textbf{69.50} \\
    & EST (5) & MLP & 63.87 & 14.61 & \textbf{70.74} \\
    & EST (1) & MLP & 65.84 & 9.47  & \textbf{67.65}\\
    & Voxel grid (10) & MLP & 52.62 & 45.23 & \textbf{80.23} \\
    & Two-channel (1) & MLP & \textbf{69.87} & 5.36 &  69.48 \\
    & EST (10) & Trilinear & 41.42 & 30.11 & \textbf{69.78} \\
    & Two-channel (1) & Trilinear & \textbf{67.84} & 6.12 & 67.42 \\
    \bottomrule
\end{tabular}
\caption{The attack success rate(\%) for shifting original events, generating additional adversarial events, and a combined method. The number in parentheses means the number of time channels in the grid representations.}
\label{table1}
\end{table*}
\section{Experiments}
In this section, we first present an extensive evaluation of adversarial attacks for event-based deep learning. We validated our algorithm with various grid representations and kernel functions on the standard event camera benchmark. All the testing results are obtained with an average of three random seeds.
\subsection{Experiments Setup}
We mainly focus on the EST model \cite{c:e5} for the attacking target: MLP kernel, EST representation, and ResNet-34 \cite{c:d2} architecture. The MLP kernel consists of two hidden layers each with 30 units. The MLP kernel takes event times as input and derives the meaningful signal around it. The temporal bins in the EST are set to 10, 5, and 1.

We follow the settings in the original EST model \cite{c:e5} to train target models: ADAM optimizer \cite{c:d5} with an initial learning rate of 0.0001 that decays by 0.5 times every 1 epoch; weight decay of 0; the  batch normalization momentum \cite{c:d6} of $0.1$. We train the networks for 30 epochs for the event camera dataset.
\subsection{Dataset}
We use N-Caltech101 dataset \cite{r:d1} in our evaluation. N-Caltech101 is the event-based version of Caltech101 \cite{c:d4}. It was recorded with an ATIS event camera \cite{r:e9} on a motor. The event camera records the event from Caltech101 examples, while the motor is moving. N-Caltech101 consists of 4,356 training samples and 2,612 validating samples in 100 classes. 
\begin{figure}[t]
\centering
\includegraphics[width=0.95\columnwidth]{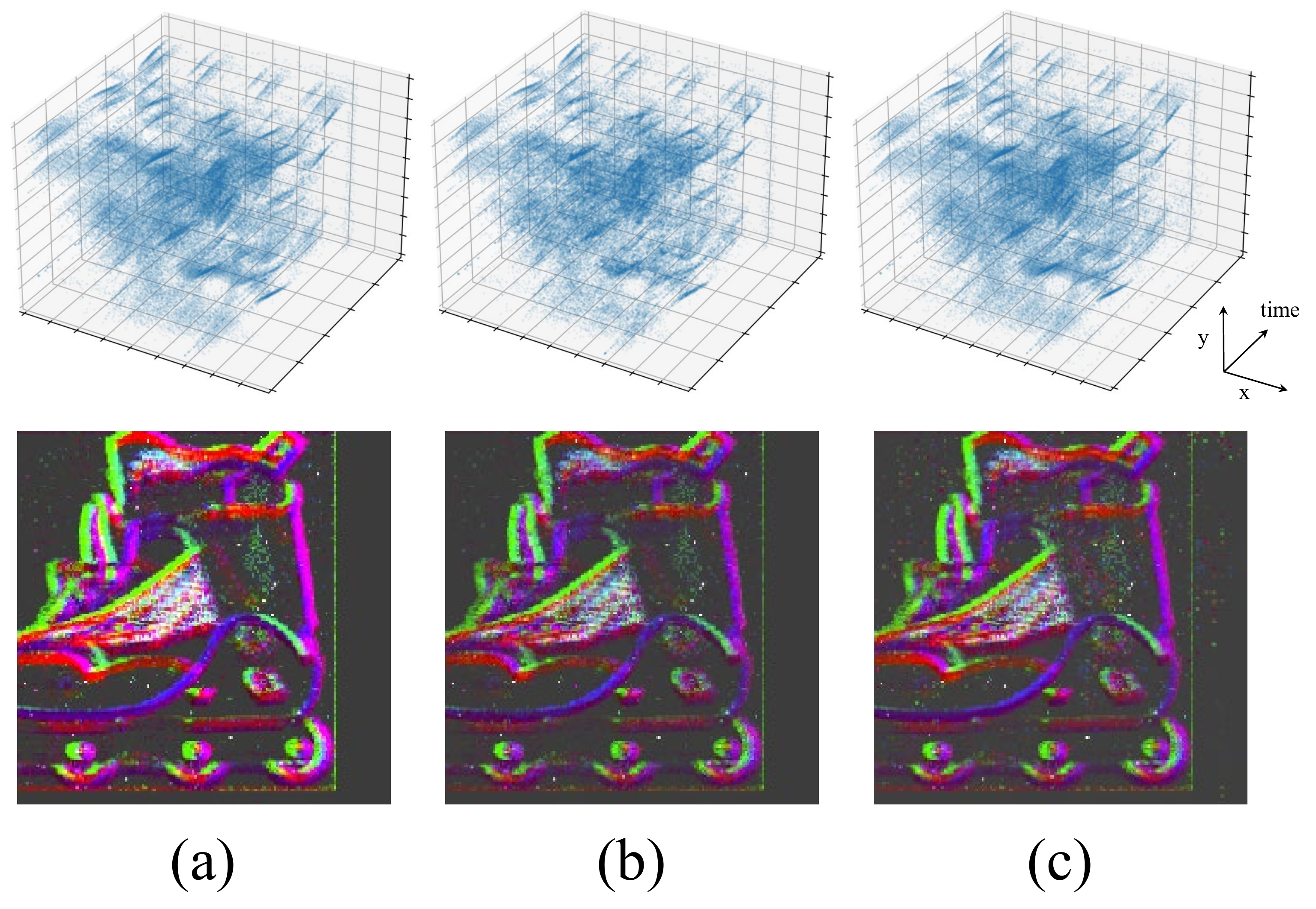} 
\caption{Visualizations of adversarial events with point-sets and grid representations. (a) Original events. (b) Shifting original events. (c) Generating additional adversarial events.}
\label{fig6}
\end{figure}
\subsection{Adversarial Attack on the Event-based Data}
In this subsection, we evaluate the attack success rate of three methods for generating adversarial examples: shifting original events, generating additional adversarial events, and a combined method. The times of the original events are shifted with a PGD attack of $L_{\infty}$ norm. We set the number of iterations to $3$. We set the default attacker step size $\alpha$ to $0.5$ and divided $\alpha$ by the number of temporal bins $B$. The $\frac{\alpha}{B}$ is the attacker step size relative to the temporal scale. But the $\frac{\alpha}{B}$ is limited below $0.1$ for visual imperceptibility. The perturbation size is set to twice of the $\frac{\alpha}{B}$. When generating additional adversarial events, we selected only the top $1\%$ loss gradients of null events and assigned five null events to the location. The number of null events is related with the visual perceptibility of adversarial examples, but the attacker parameters of generating additional adversarial events had little effect on the perceptibility. Therefore, we set the attacker parameters larger compared to the shifting events: attacker step size to $0.01$, the perturbation size to $0.1$, and the number of iterations to $10$. 

The attacking targets are EST, voxel grid, and two-channel representations models. EST (10), EST (5), and EST (1) have ten, five, and one temporal bin(s) for each polarity, respectively. Two-channel (1) has one temporal bin as EST (1), but it is derived by averaging the EST (10) temporal bins. Voxel grid (10) is derived by averaging the EST (10) polarity channels.

As shown in Table \ref{table1}, shifting original events substantially reduces the model's recognition performance because it distorts the features of the original events. This method achieves a 95.50\% untargeted attack success rate with a small step size of $0.05$ for the EST (10). However, larger step size of $0.1$ is required to fool the EST (1) and two-channel (1). Generating additional adversarial events does not significantly reduce the recognition performance compared to shifting events. However, additional adversarial events improve the attack success rate of shifting events for most models.

Figure \ref{fig6} shows the adversarial events of the EST (10) from Table \ref{table1}. Shifting events changes the times of the original events, resulting in the color change in grid representation. Additional adversarial events create adversarial features in the grid representation. Visualization results show that all the attack method fool the deep learning model with imperceptible changes. More visualization results can be found in the supplemental material.
\subsection{The Effect of Perturbation Size}
\begin{table}[t]
\small
\centering
\begin{tabular}[t]{c|ccccc}
    \toprule
    \diagbox[width=8em]{Model}{$\epsilon$} & $0.05$ & $0.10$ & $0.15$ & $0.2$ & $0.25$\\
    \midrule
    EST (10) & 95.11 & \textbf{95.50} & 93.40 & 94.96 & 90.66 \\
    EST (5) & 90.56 & 94.74 & 95.49 & \textbf{96.13} & 94.83\\
    EST (1) & 62.61 & 86.01 & 93.63 & 95.96 & \textbf{98.05}\\
    Voxel grid (10) & {94.89} & \textbf{95.14} & 93.18 & 89.86 & 87.99\\
    Two-channel (1) & 69.77 & 89.72 & 95.8 & 97.73 & \textbf{98.31}\\
    \bottomrule
\end{tabular}
\caption{The untargeted attack success rate (\%) of shifting events against EST (10), EST (5), EST (1), voxel grid (10), and two-channel (1) with respect to the perturbation size $\epsilon$.}
\label{table2}
\end{table}
\begin{table}[t]
\small
\centering
\begin{tabular}[h]{c|cccccc}
    \toprule
    \diagbox[width=8em]{Frequency}{$\epsilon$} & $0.05$ & $0.10$ & $0.15$ & $0.2$ & $0.25$\\
    \midrule
    Normal & 95.11 & \textbf{95.50} & 93.40 & 94.96 & 90.66 \\
    Half & 75.5 & 86.17 & 90.18 & \textbf{91.16} & 81.79 \\
    \bottomrule
\end{tabular}
\caption{The untargeted attack success rate (\%) of shifting events against EST (10) with respect to the frequency of data and perturbation size $\epsilon$.}
\label{table_frequency}
\end{table}
Large perturbations increase the attack success rate in general. In the adversarial attack on the event-based data, perturbation size relative to the temporal scale is more important. Temporal scale is determined by the temporal bins of grid representation model and the data frequency. 

Tables \ref{table2} and \ref{table_frequency} show the attack success rates for shifting events with different perturbation sizes. The step size is set to half of the perturbation size, and the number of iterations is set to three. As shown in Table \ref{table2}, the attack success rates of the EST (10), EST (5), and voxel grid (10) do not increase continuously as the perturbation size increases. The attack success rate of EST (10) and voxel grid (10) is oscillating. For example, the optimal perturbation size is $0.05$ for attacking the EST (10) model. EST (1) and two-channel (1) have only one temporal bins, thus the attack success rates increase continuously.

Table \ref{table_frequency} shows the attack success rates for the EST (10) with respect to data frequency and perturbation size. Data frequency is determined by the repeated movements of an event camera. We decrease the data frequency by half by using the first half of the time and normalizing the time to one. Our results show that the perturbation size is inversely proportional to the number of temporal bins $B$ and data frequency. 
\subsection{Adversarial Training on the Event-based Data}
\begin{table}[t]
\small
\centering
%\resizebox{.95\columnwidth}{!}{
\begin{tabular}[t]{c|cccc}
    \toprule
     \diagbox[width=8em]{Training}{Attack} & None & Shifting & Shifting \& generating\\
    \midrule
    Original & 84.99 & 2.99 & 0 \\
    Adversarial & 80.67 & 52.79 & 22.86 \\
    \bottomrule
\end{tabular}
\caption{The top-1 accuracy (\%) of the EST (5) model against untargeted attacks. The adversarial training model is trained with original events and adversarial events.}
\label{table4}
\end{table}
In this subsection, we use the proposed attack algorithm to train networks on the perturbed features. The goal of this adversarial training is to produce networks that perform well both on the original events and adversarial events. We started on the pretrained model and trained the models for 5 epochs with ADAM optimizer with initial learning rate of $0.00001$. Each model is trained with original events and adversarial events. Table \ref{table4} shows the robustness of each model to the attack methods. The results demonstrate that the adversarial training model improves the robustness on the adversarial events compared to the original model.
\subsection{Transferability of Adversarial Events}
\begin{table}[t]
\centering
\small
%\resizebox{.95\columnwidth}{!}{
\begin{tabular}[t]{c|ccc}
    \toprule
     \diagbox[width=10em]{To}{From} & EST (10) & Voxel grid (10) & EST (1)  \\
    \midrule
    EST (10) & - & 27.51  &  27.51  \\
    Voxel grid (10) & 27.47 & - & 46.85 \\
    EST (1) & 0.72 & 0.86 & -  \\
    \bottomrule
\end{tabular}
\caption{The attack success rate (\%) of untargeted transfer attacks against EST (10), voxel grid (10), and EST (1).}
\label{table5}
\end{table}
We feed adversarial events of EST (10), EST (1), and voxel grid (10) models to other models. Each model is trained with different weight initialization. % Adversarial events are generated by shifting events with attacker step size of $0.05$. The attacker step size is set based on the high attack success rate from Table \ref{table2}.
Table \ref{table5} shows that adversarial events of EST (1) transfer to other models. However, adversarial events of EST (10) and voxel grid (10) hardly transfer to the EST (1). As the EST (10) and voxel grid (10) have smaller temporal scale in each temporal bin, the adversarial events of each temporal bin are merged into the larger temporal scale of the EST (1). Thus, the perturbations are eliminated in the EST (1). On the contrary, adversarial events of EST (1) perturbs the EST (10) and voxel grid (10) along all temporal bins without elimination. 
\section{Conclusion}
To the best of our knowledge, we generated adversarial events to fool the event-based deep learning models for the first time. Our attack algorithm shifts the original event times and generates additional adversarial events. For shifting events, we set the perturbation size relative to the data frequency and temporal scales of the model. Additional adversarial events boost the attack performance for most models. Our experimental results show that the proposed algorithm can find adversarial events with an average of 97.95\% for untargeted attack (refer to Table \ref{table1}). We hope this work can provide a guideline for future adversarial event research. 
\section{Acknowledgements}
The student is supported by the BK21 FOUR from the Ministry of Education (Republic of Korea).
\bibliography{aaai22}

\end{document}